\documentclass{article}
\usepackage{spconf,amsmath,graphicx}
\usepackage[T1]{fontenc}
\usepackage[linesnumbered,ruled]{algorithm2e}
\usepackage{stfloats,color}
\usepackage{amsfonts}
\usepackage{amssymb}
\usepackage{cite}
\usepackage{array}
\usepackage{subcaption}
\usepackage{times}
\usepackage{epsfig}
\usepackage{latexsym}
\usepackage{epstopdf}
\usepackage{verbatim}
\usepackage{units}
\usepackage{amsthm}
\usepackage{placeins}
\usepackage{afterpage}
\usepackage{dsfont}
\usepackage{soul}
\usepackage{multicol}
\usepackage{multirow}
\usepackage{mathtools}
\usepackage[cmintegrals]{newtxmath}
\usepackage{url}
\newcolumntype{P}[1]{>{\centering\arraybackslash}p{#1}}
\newcolumntype{M}[1]{>{\centering\arraybackslash}m{#1}}

\newcommand{\defeq}{\ensuremath{\triangleq}}

\title{Distributed Gradient Descent with \\ Coded Partial Gradient Computations}
\name{E. Ozfatura$^\dagger$, S. Ulukus$^+$ and D. G{\"u}nd{\"u}z$^\dagger$}
\address{$^\dagger$ Department of Electrical and Electronic Engineering, Imperial College London, UK \\
$^+$ Department of Electrical and Computer Engineering, 
University of Maryland, MD }

\begin{document}
%
\maketitle
\begin{abstract}
Coded computation techniques provide robustness against \textit{straggling} servers in distributed computing, with the following limitations: First, they increase decoding complexity. Second, they ignore computations carried out by straggling servers; and they are typically designed to recover the full gradient, and thus, cannot provide a balance between the accuracy of the gradient and per-iteration completion time. Here we introduce a hybrid approach, called \textit{coded partial gradient computation (CPGC)}, that benefits from the advantages of both coded and uncoded computation schemes, and reduces both the computation time and decoding complexity. 
\end{abstract}
\section{Introduction}
\label{sec:intro}
In many machine learning applications, the principal computational task boils down to a matrix-vector multiplication. Consider, for example, the minimization of the empirical mean squared error in linear regression
$L(\boldsymbol{\theta}) \triangleq \frac{1}{2N}\sum_{i=1}^{N}(y_{i}-\mathbf{x}_{i}^{T}\boldsymbol{\theta})^{2}$, where $\mathbf{x}_{1},\ldots,\mathbf{x}_{N} \in \mathbb{R}^{L}$
are the data points with the corresponding labels $y_{1},\ldots,y_{N} \in \mathbb{R}$, and $\boldsymbol{\theta}\in \mathbb{R}^{L}$ is the parameter vector. The optimal parameter vector can be obtained iteratively by gradient descent (GD): $\boldsymbol{\theta}_{t+1}=\boldsymbol{\theta}_{t}-\eta_{t} \nabla_{\boldsymbol{\theta}} L(\boldsymbol{\theta}_{t})$,
 where $\eta_{t}$ is the learning rate and $\boldsymbol{\theta}_{t}$ is the parameter vector at the $t$th iteration. We have $\nabla_{\boldsymbol{\theta}} L(\boldsymbol{\theta_t}) = \mathbf{X}^{T} \mathbf{X} \boldsymbol{\theta}_{t}-\mathbf{X}^{T}\mathbf{y}$, where $\mathbf{X}=[\mathbf{x}_{1},\ldots,\mathbf{x}_{N}]^{T}$ and 
$\mathbf{y}=[y_{1},\ldots,y_{N}]^{T}$. In the gradient expression, only $\boldsymbol{\theta}_{t}$ changes over the iterations; hence, the key computational task at each iteration is the  matrix-vector multiplication $\mathbf{W}\boldsymbol{\theta}_{t}$, where $\mathbf{W}\defeq\mathbf{X}^{T}\mathbf{X}\in\mathbb{R}^{L\times L}$. To speed up GD, execution of this multiplication can be distributed to $K$ \textit{worker} servers, by simply dividing $\mathbf{W}$ into $K$ equal-size disjoint submatrices. However, the computation time will now be limited by the {\em straggling} workers.\\
\indent Coded distributed computation has been introduced to tolerate straggling workers by introducing redundant computations \cite{CC.1,CC.2, CC.3,CC.4,CC.5,CC.6,CC.7,UNPS,UNPS2}. Maximum distance separable (MDS) codes are used in \cite{CC.1}, where  matrix $\mathbf{W}\in\mathbb{R}^{L\times L}$  is divided into $M$ disjoint submatrices, $\mathbf{W}_{1},\ldots,\mathbf{W}_{M}\in\mathbb{R}^{r\times L},$  which are then encoded with an $(M,K)$ MDS code, and each coded submatrix is assigned to a different worker. Each worker multiplies $\boldsymbol{\theta}_t$ with the coded  submatrix assigned to it, and sends the result to the master, which can recover $\mathbf{W}\boldsymbol{\theta}_{t}$ 
having received the results from any $M$ workers. Up to $K-M$ stragglers can be tolerated at the expense of increasing the \textit{computation load} of each worker by $r = L/M$ \cite{CC.1}. Alternatively,  uncoded computations can be executed, and  the results can be send as a coded messages \cite{UCCT.1,UCCT.2,UCCT.3}. However, these approaches completely discard computations carried out by straggling servers, and hence, the overall computational capacity is underutilized.

Alternatively, workers can be allowed to send multiple messages to the master per-iteration, corresponding to partial computations \cite{UNPS, CC.2, CC.5, UCUT.4}, which will be called {\em multi-message communication (MMC)}. In \cite{CC.2} MMC is applied to MDS-coded computation utilizing the statistics of stragglers. Instead, rateless codes are proposed in \cite{UNPS} as they do not require the knowledge of the straggler statistics, and also reduce the decoding complexity. However, rateless codes come with an overhead, which vanishes only if the number of codewords goes to infinity. This, in turn, would increase the number of read/write operations at the master at each iteration, limiting the practicality in real applications. 

Uncoded distributed computation with MMC (UC-MMC) is introduced in \cite{UCUT.4,UCUT.2,CC.5}, and is shown to outperform coded computation in terms of average completion time, concluding that coded computation is more effective against persistent stragglers, and particularly when full gradient is required at each iteration. Coded GD strategies are mainly designed for full gradient computation; and hence, the master needs to wait until all the gradients can be recovered. UC-MMC, on the other hand, in addition to exploiting partial computations performed by straggling servers, also allows the master to update the parameter vector with only a subset of the gradient computations to limit the per iteration completion time.

In this paper, we introduce a novel hybrid scheme, called \textit{coded partial gradient computation (CPGC)}, that brings together the advantages of uncoded computation, such as low decoding complexity and partial gradient updates, with those of coded computation, such as reduced per-iteration completion time and limited communication load. Before presenting the design principles of this scheme, we will briefly outline its advantages on a simple motivating example.

\begin{table}[t]{\footnotesize
    \begin{center}
    \begin{tabular}{ | p{3.5cm} | p{0.6cm} | p{1.2cm} |p{0.6cm}|}
    \hline
    cumulative computation type&   MCC &  UC-MMC& CPGC\\ \hline
    $\mathbf{N}_{1}:N_{2}=4,N_{1}=0,N_{0}=0$ &1 & 1 &1 \\ \hline $\mathbf{N}_{2}:N_{2}=3,N_{1}=1,N_{0}=0$ &4 &4 &4 \\ \hline $\mathbf{N}_{3}:N_{2}=3,N_{1}=0,N_{0}=1$ & 4& 4 &4 \\ \hline $\mathbf{N}_{4}:N_{2}=2,N_{1}=2,N_{0}=0$ &6&6 &6 \\ \hline $\mathbf{N}_{5}:N_{2}=2,N_{1}=1,N_{0}=1$ & 12& 8 &12 \\ \hline $\mathbf{N}_{6}:N_{2}=2,N_{1}=0,N_{0}=2$ & 6&2 &6 \\ \hline
  $\mathbf{N}_{7}:N_{2}=1,N_{1}=3,N_{0}=0$ & 0&4 &4 \\ \hline $\mathbf{N}_{8}:N_{2}=1,N_{1}=2,N_{0}=1$ & 0& 4 &8 \\ \hline
$\mathbf{N}_{9}:N_{2}=0,N_{1}=4,N_{0}=1$ & 0& 1 &1 \\ \hline	
		\end{tabular}
		\caption{Number of  score vectors for full gradient.}\label{table: ex1full}	          
		\end{center}
		       }
		       \vspace*{-0.8cm}
\end{table}

\section{Motivating Example}
\label{sec:mot}

Consider $M=4$ computation tasks, represented by submatrices $\mathbf{W}_1, \ldots, \mathbf{W}_4$, which are to be executed across $K=4$ workers, each with a maximum computation load of $r=2$; that is, each worker can perform up to $2$ computations, due to storage or computation capacity limitations. Let us first consider two known distributed computation schemes, namely UC-MMC \cite{UCUT.4, CC.5} and MDS-coded computation (MCC) \cite{CC.1}. 

For each scheme, the $r \times K$ \textit{computation  scheduling matrix}, $\mathbf{A}$, shows the assigned computation tasks to each worker with their execution order. More specifically,  $\mathbf{A}(i,j)$ denotes the $i$th computation task to be executed by the $j$th worker. 
In MCC, linearly independent coded computation tasks are distributed to the workers as follows:
\[
\mathbf{A}_{m}=
  \begin{bmatrix}
    \mathbf{W}_{1}+ \mathbf{W}_{3} &  \mathbf{W}_{1}+2\mathbf{W}_{3} &   \mathbf{W}_{1}+4\mathbf{W}_{3} &  \mathbf{W}_{1}+8\mathbf{W}_{3} &  \\
    \mathbf{W}_{2}+\mathbf{W}_{4} &  \mathbf{W}_{2}+2\mathbf{W}_{4} &  \mathbf{W}_{2}+4\mathbf{W}_{4} &  \mathbf{W}_{2}+8\mathbf{W}_{4} & \\  
  \end{bmatrix}.
\]
Each worker sends the results of its computations only after  all of them are completed,  i.e., first worker sends the concatenation $[(\mathbf{W}_{1} + \mathbf{W}_{3})\boldsymbol{\theta}_t ~ \text{ }(\mathbf{W}_{2}+\mathbf{W}_{4})\boldsymbol{\theta}_t]$ after completing both computations; therefore, any permutations of each column vector would result in the same performance. $\mathbf{A}_{m}$ corresponds to a $(2,4)$ MDS code, and hence, the master can recover the full gradient computation from the results of any two workers.

In the UC-MMC scheme with a shifted computation schedule\cite{CC.5}, computation scheduling matrix is given by
\[
\mathbf{A}_{u}=
  \begin{bmatrix}
    \mathbf{W}_{1} & \mathbf{W}_{2} & \mathbf{W}_{3} & \mathbf{W}_{4} \\
   \mathbf{W}_{2} & \mathbf{W}_{3}& \mathbf{W}_{4}  & \mathbf{W}_{1} \\
  \end{bmatrix},
\]
and each worker sends the results of its computations sequentially, as soon as each of them is completed. This helps to reduce the per-iteration completion time with an increase in the communication load \cite{UCUT.4, CC.5}.  With UC-MMC, full gradient can be recovered even if each worker performs only one computation, which is faster if the workers have similar speeds.

The computation scheduling matrix of CPGC is given by
 \[
\mathbf{A}_{c}=
  \begin{bmatrix}
    \mathbf{W}_{1} & \mathbf{W}_{2} & \mathbf{W}_{3} & \mathbf{W}_{4} \\
    \mathbf{W}_{3}+\mathbf{W}_{4}& \mathbf{W}_{1}+\mathbf{W}_{3} & \mathbf{W}_{2}+\mathbf{W}_{4} & \mathbf{W}_{1}+\mathbf{W}_{2}  \\  
  \end{bmatrix}.
  \]
  

\subsection{Full Gradient Performance}

Now, let us focus on a particular iteration, and let $N_{s}$ denote the number of workers that have completed exactly $s$  computations by time $t$, $s = 0, \ldots, r$. We define  $\mathbf{N} \triangleq (N_{0},\ldots,N_{r})$ as the {\em cumulative computation type}. Additionally, we introduce the $K$-dimensional \textit{score vector} $\mathbf{C}=[c_{1},\ldots,c_{K}]$, where $c_{i}$ denotes the number of computations completed by the $i$th worker. For each scheme, the number of distinct score vectors with the same cumulative computation type, which allow the recovery of full gradient is listed in Table \ref{table: ex1full}. Particularly striking are the last three rows that correspond to cases with very few computations completed, i.e., when at most one worker completes all its assigned tasks. In these cases, CPGC is much more likely to allow full gradient computation; and hence, the computation deadline can be reduced significantly while still recovering the full gradient.

\begin{table}[t]{\footnotesize
    \begin{center}
    \begin{tabular}{ | p{3.5cm} | p{0.6cm} | p{1.2cm} |p{0.6cm}|}
    \hline
    cumulative computation type & MCC & UC-MMC& CPGC\\ \hline
    $\mathbf{N}_{1}:N_{2}=4,N_{1}=0,N_{0}=0$ &1 & 1 &1 \\ \hline $\mathbf{N}_{2}:N_{2}=3,N_{1}=1,N_{0}=0$ &4 &4 &4 \\ \hline $\mathbf{N}_{3}:N_{2}=3,N_{1}=0,N_{0}=1$ & 4& 4 &4 \\ \hline $\mathbf{N}_{4}:N_{2}=2,N_{1}=2,N_{0}=0$ &6&6 &6 \\ \hline $\mathbf{N}_{5}:N_{2}=2,N_{1}=1,N_{0}=1$ & 12& 12 &12 \\ \hline $\mathbf{N}_{6}:N_{2}=2,N_{1}=0,N_{0}=2$ & 6&6 &6 \\ \hline
  $\mathbf{N}_{7}:N_{2}=1,N_{1}=3,N_{0}=0$ & 0&4 &4 \\ \hline $\mathbf{N}_{8}:N_{2}=1,N_{1}=2,N_{0}=1$ & 0& 12 &12 \\ \hline
 $\mathbf{N}_{9}:N_{2}=1,N_{1}=1,N_{0}=2$ & 0& 8 &8 \\ \hline
 $\mathbf{N}_{10}:N_{2}=0,N_{1}=4,N_{0}=0$ & 0& 1 &1 \\ \hline
 $\mathbf{N}_{11}:N_{2}=0,N_{1}=3,N_{0}=1$ & 0& 4 &4 \\ \hline
		\end{tabular}
		\caption{Number of  score vectors for partial gradient.}\label{table: ex1par}	          
		\end{center}
		       }
		       \vspace*{-0.8cm}
\end{table}

Next, we analyze the  probability of each type under a specific computation time statistics. We adopt the model in \cite{comptime}, where the  probability of completing exactly $s$  computations by time $t$, $P_{s}(t)$, is given by
\begin{equation}\label{stat}
P_{s}(t)= 
    \begin{cases}
     0, &  \text{if } t<s\alpha, \\
     1-  e^{-\mu(\frac{t}{s}-\alpha)} ,  & s\alpha \leq t <(s+1) \alpha,\\
           e^{-\mu(\frac{t}{s+1}-\alpha)}-e^{-\mu(\frac{t}{s}-\alpha)}   &(s+1)\alpha<t, 
     \end{cases}
\end{equation}
where  $\alpha$ is the minimum required time to finish a computation task, and $\mu$ is the average number of computations completed in unit time. The probability of cumulative computation type $\mathbf{N}(t)$ at time $t$ is given by $\mathrm{Pr}(\mathbf{N}(t))=\prod_{s=0}^{r} P_{s}(t)^{N_{s}}$.
Let $T$ denote the full gradient recovery time. Accordingly, $\mathrm{Pr}(T<t)$ for CPGC is given by
\begin{align}\label{pc}
&\mathrm{Pr}(\mathbf{N}_{1}(t))+4\mathrm{Pr}(\mathbf{N}_{2}(t))+4\mathrm{Pr}(\mathbf{N}_{3}(t))+6\mathrm{Pr}(\mathbf{N}_{4}(t))\nonumber+12\mathrm{Pr}(\mathbf{N}_{5}(t))\\&+6\mathrm{Pr}(\mathbf{N}_{6}(t))+4\mathrm{Pr}(\mathbf{N}_{7}(t))+8\mathrm{Pr}(\mathbf{N}_{8}(t))+\mathrm{Pr}(\mathbf{N}_{9}(t))
\end{align}
where the types $\mathbf{N}_{1}, \ldots, \mathbf{N}_{9}$ are as listed in Table \ref{table: ex1full}. $\mathrm{Pr}(T<t)$ for MCC and UC-MMC can be  written similarly. Then, one can observe that, for any $t$, CPGC has the highest $\mathrm{Pr}(T<t)$; and hence, the minimum average per-iteration completion time $E[T]$.  In the next subsection, we will highlight the partial recoverability property of CPGC.    

\subsection{Partial Gradient Performance}
It is known that stochastic GD can still guarantee convergence even if each iteration is completed with only a subset of the gradient computations \cite{UCUT.1,SGD1}. In our example, with three out of four gradients, sufficient accuracy may be achieved at each iteration, particularly if the straggling server is varying over iterations. The number of score vectors for which a partial gradient (with at least three gradient computations) can be recovered are given in Table \ref{table: ex1par}. We observe that when three gradients are sufficient to complete an iteration UC-MMC and CPGC have the same average completion time statistics. Hence, CPGC can provide a lower average per-iteration completion time for full gradient computation compared to UC-MMC, while achieving  the same performance when partial gradients are allowed. 

\section{Design Principles of CPGC}\label{sec:Design}

In \cite{UNPS}, LT codes are proposed for distributed computation in order to exploit MMC with coded computations. However, LT codes come with a trade-off between the overhead and the associated coding/decoding complexity. Moreover, the original design in \cite{UNPS} does not allow partial gradient recovery. 




The key design issue in an LT code is the degree distribution $P(d)$. Degree of a codeword, $d$, chosen randomly from $P(d)$, defines the number of symbols ($\mathbf{W}_i$ submatrices in our setting) that are used in generating a codeword. Then, $d$ symbols are chosen randomly to form a codeword. The degree distribution plays an important role in the performance of an LT code, and the main challenge is to find the optimal degree distribution. Codewords with smaller degrees reduce decoding complexity; however, having many codewords with smaller degrees increases the probability of linear dependence among codewords. We also note that, LT code design is based on the assumption that the erasure probability of different codewords are identical and independent from each other. However, in a coded computing scenario, the computational tasks, each of which corresponding to a distinct codeword, are executed sequentially; thus, erasure probabilities of codewords are neither identical nor independent. Codewords must be designed taking into account their execution orders in order to prevent overlaps and to minimize the average completion time. This is the main intuition behind the CPGC scheme, and guides the design of the computation scheduling matrix. 

\subsection{Degree Limitation}
To allow partial gradient computation at the master, we limit the degree of all codewords by two; that is, each codeword (i.e., coded submatrix) is the sum of at most two submatrices. Moreover, the first computation task assigned to each worker corresponds to a codeword with degree one (i.e., a $\mathbf{W}_i$ submatrix is assigned to each worker without any coding), while all other tasks correspond to codewords with degree two (coded submatrices). Recall that, due to the straggling behavior, the first task at each worker has the highest completion probability, thus assigning uncoded submatrices as the first computation task at each worker helps to enable partial recovery.

\subsection{Coded Data Generation}
In an LT code, symbols (submatrices) that are linearly combined to generate a codeword are chosen randomly; however, to enable partial gradient recovery, we carefully design the codewords for each worker. 

For a given set of submatrices $\mathcal{W}$, a partition $\mathcal{P}$ is a grouping of its elements into nonempty disjoint subsets. In our example, we have $\mathcal{W}=\left\{\mathbf{W}_{1}, \mathbf{W}_{2}, \mathbf{W}_{3}, \mathbf{W}_{4} \right\}$, and $\mathcal{P}=\left\{\left\{\mathbf{W}_{1},\mathbf{W}_{2}\right\}, \left\{\mathbf{W}_{3}, \mathbf{W}_{4} \right\}\right\}$ is a partition. Now, consider the following scheme: for each $\mathcal{Q} \in \mathcal{P}$,  a codeword $c(\mathcal{Q})$ is generated by $\sum_{\mathbf{W'}\in \mathcal{Q}} \mathbf{W'}$. Since for any $\mathcal{Q}_{i},\mathcal{Q}_{j}\in \mathcal{P}$, $i \neq j$, $\mathcal{Q}_{i}\cap \mathcal{Q}_{j}=\emptyset$, codewords $c(\mathcal{Q}_{i})$ and $c(\mathcal{Q}_{j})$ share no common submatrix. Accordingly, one can easily observe that if $n$ partitions are used to generate coded submatrices, each submatrix $\mathbf{W}_{i}$ appears in exactly $n$ different coded submatrices. In order to generate degree-two codewords, we use  partitions with subsets of size two; and hence, exactly $K/2$ coded submatrices are generated from a single partition. Therefore, for each row of the computation scheduling matrix we need exactly two partitions of $\mathcal{W}$, and in total we require $2(r-1)$ distinct partitions (see \cite{long} for details).

Note that the probability of not receiving the results of computations corresponding to coded submatrices in the same column of the computation scheduling matrix are correlated, as they are executed by the same worker. Hence, in order to minimize the dependence on a single worker, we would like to limit the appearance of a submatrix in any single column of the computation scheduling matrix. In the next section, we provide a heuristic strategy for coded submatrix assignment.

\section{Numerical Results  and Conclusions}
We will analyze and compare the performance of three schemes, UC-MMC, CPGC and MCC, in terms of three performance measures, the \textit{average per-iteration completion time}, \textit{communication load} and the \textit{communication volume}. The communication load, defined in \cite{UCUT.4,CC.5}, refers to the average number of messages transmitted to the master from the workers per iteration, whereas the communication volume refers to the average total size of the computations sent to the master per iteration. This is normalized with respect to the result of $\mathbf{W}\mathbf{\theta}$, which is set as the unit data volume. This is to distinguish between the partial and full computation results sent from the workers in CPGC and MCC schemes, respectively. In CPGC we transmit many messages of smaller size, while MCC sends a single message consisting of multiple results. Communication volume allows us to compare the amount of redundant computations sent from the workers to the master. A communication volume of $1$ implies zero communication overhead, whereas a communication volume larger than $1$ implies communication overhead due to transmission of multiple messages. 
\begin{figure*}
    \centering
         \begin{subfigure}[b]{0.33\textwidth}
        \includegraphics[scale=0.4]{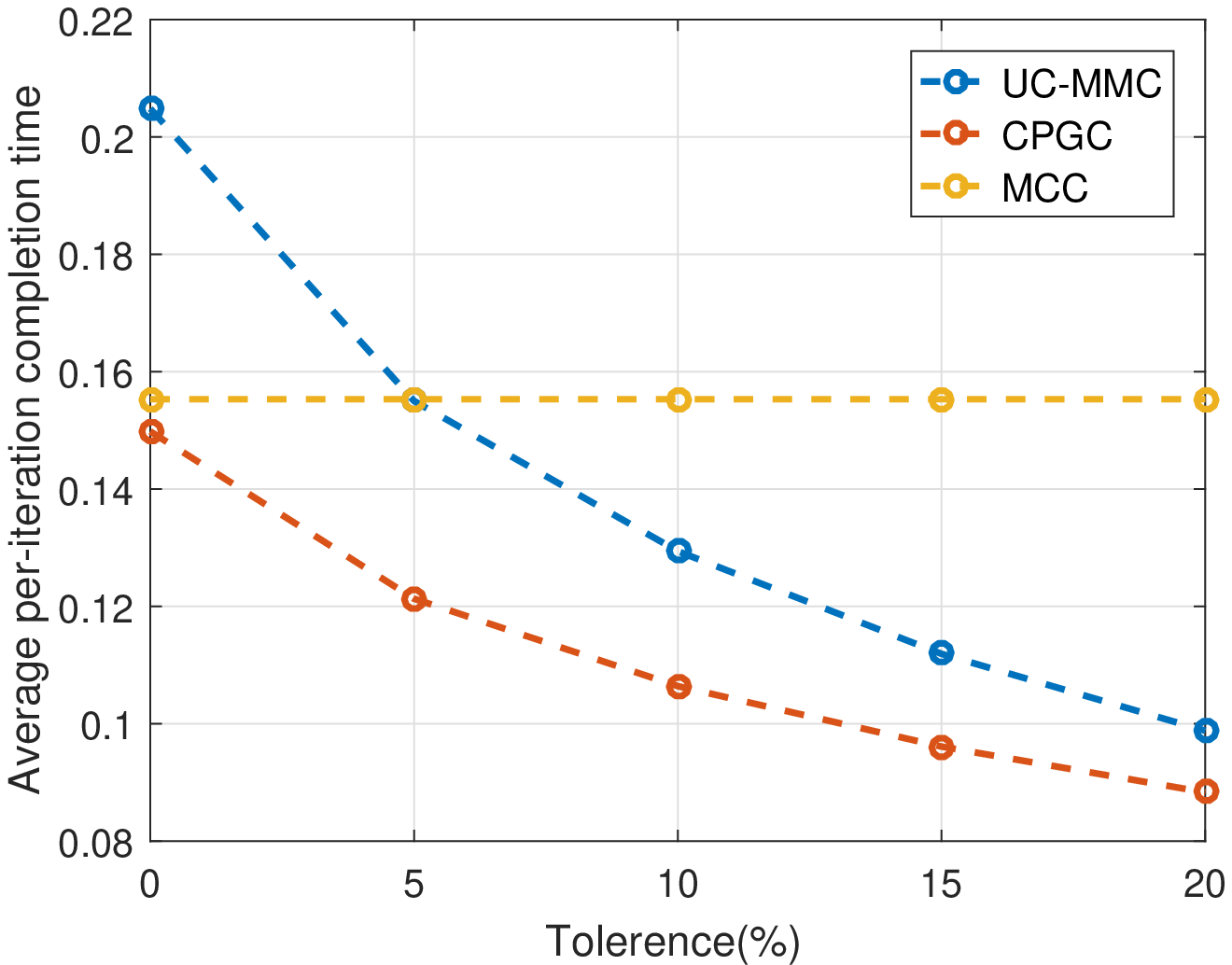}
        \caption{Average per-iteration time comparison.}
				\label{comp1}
    \end{subfigure}
    \begin{subfigure}[b]{0.33\textwidth}
        \includegraphics[scale=0.4]{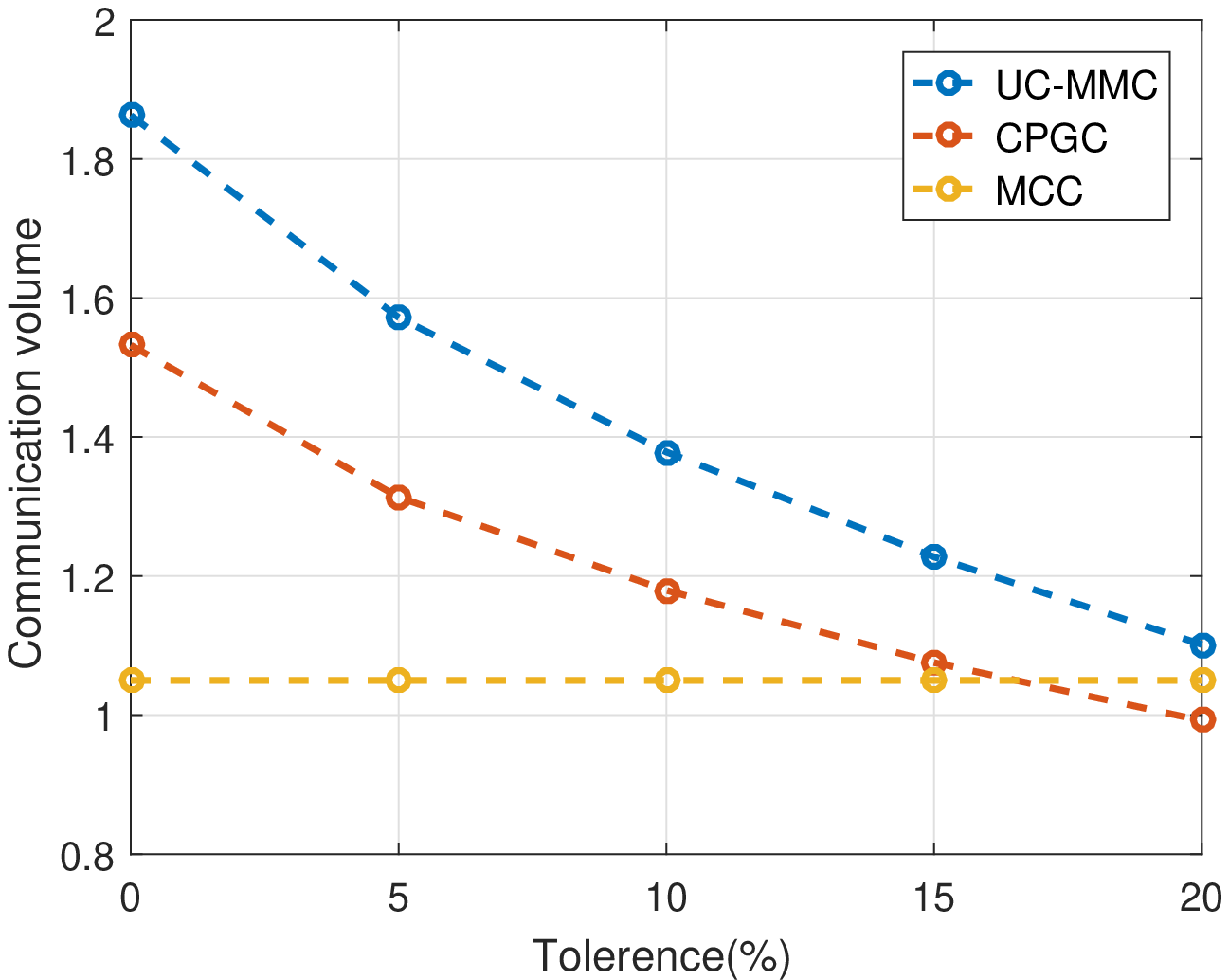}
        \caption{Communication volume comparison.}
				\label{comm1}
        \end{subfigure}
        \begin{subfigure}[b]{0.33\textwidth}
        \includegraphics[scale=0.4]{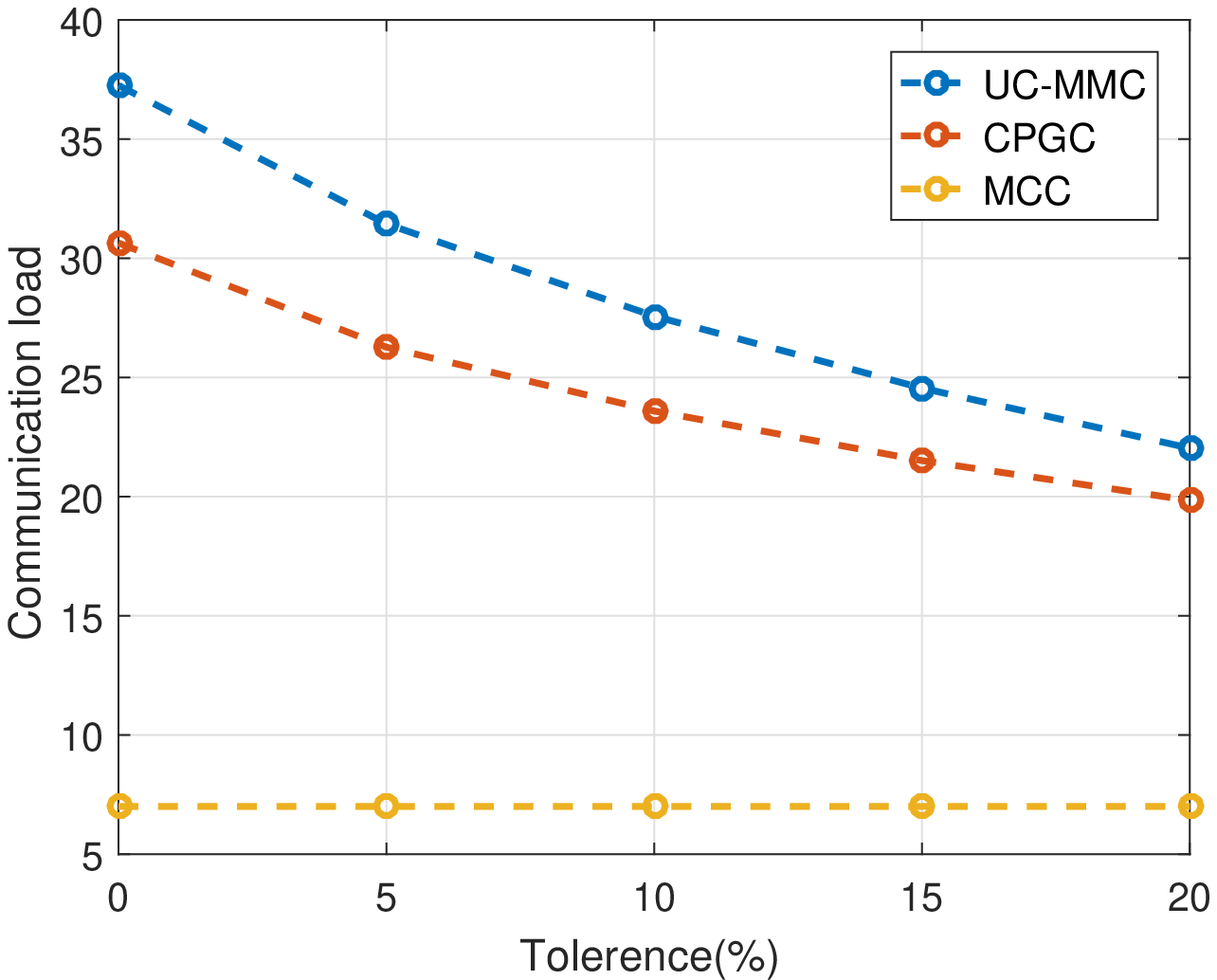}
        \caption{Communication load comparison.}
				\label{pack1}
        \end{subfigure}
				\caption{Performance comparison of UC-MCC, CPGC and MCC schemes for $M=K=20$ and $r=3$}
		\label{sim1}
		\vspace*{-0.4cm}
\end{figure*}

\subsection{Simulation Setup}
We consider $K=20$ workers and $M=20$ computation tasks (submatrices), and a computation load of $r=3$. We set $\mu=10$ and $\alpha=0.01$ for the statistics of computation speed in (\ref{stat}). 

In CPGC, first computations assigned to the workers are uncoded submatrices. For the second and third rows of the computation scheduling matrix we use four different partitions with the coded submatrices as follows (assuming $N$ is even):
\begin{align} \label{partset}
\mathbf{v}_{1}=&[\mathbf{W}_{1}+\mathbf{W}_{2},\ldots,\mathbf{W}_{n}+\mathbf{W}_{n+1} ,\ldots,\mathbf{W}_{N-1}+\mathbf{W}_{N}]\nonumber\\
\mathbf{v}_{2}=&[\mathbf{W}_{1}+\mathbf{W}_{3},\ldots,\mathbf{W}_{n}+\mathbf{W}_{n+2} ,\ldots,\mathbf{W}_{N-2}+\mathbf{W}_{N}]\nonumber\\
\mathbf{v}_{3}=&[\mathbf{W}_{1}+\mathbf{W}_{N},\ldots,\mathbf{W}_{n}+\mathbf{W}_{N-n+1} ,\ldots,\mathbf{W}_{N/2}+\mathbf{W}_{N/2+1}]\nonumber\\
\mathbf{v}_{4}=&[\mathbf{W}_{1}+\mathbf{W}_{N/2+1},\ldots,\mathbf{W}_{n}+\mathbf{W}_{N/2+n} ,\ldots,\mathbf{W}_{N/2}+\mathbf{W}_{N}]\nonumber
\end{align}
These coded submatrices are used to form a computation scheduling matrix in the following way: $\mathbf{A}(2, 1:K/2) = \text{circshift}(\mathbf{v}_{1};-1)$,
$\mathbf{A}(2,K/2+1:K) = \text{circshift}(\mathbf{v}_{2};-1)$,
$\mathbf{A}(3,1:K/2) = \text{circshift}(\mathbf{v}_{3};1)$,
$\mathbf{A}(3,K/2+1:K) = \text{circshift}(\mathbf{v}_{4};-2)$, where  $\text{circshift}$ is  the circular shift operator, i.e., $\text{circshift}(\mathbf{v};d)$ is the $d$ times right shifted version of vector $\mathbf{v}$. We use the shifted version of the vectors to prevent multiple appearance of a submatrix in a single column. 

\subsection{Results}
For $M$ submatrices, let  $M'$ be the required number of computations, each corresponding to a different submatrix, to terminate an iteration. We define $\frac{M-M'}{M}$ as the \textit{tolerance rate}, which reflects the gradient accuracy at each iteration (lower tolerance rate means higher accuracy).

In Fig. \ref{sim1}, we compare the three schemes under the three performance metrics with respect to the tolerance rate. Since partial recovery is not possible with MDS-coded computation, its performance remains the same with the tolerance level. The performance of the UC-MMC and CPGC schemes improve with the increasing tolerance level. This comes at the expense of a slight reduction in the accuracy of the resultant gradient computation. We remark that, beyond a certain tolerance level UC-MMC scheme achieves a lower average per iteration completion time compared to MCC due to the utilization of non-persistent stragglers thanks to the MMC approach \cite{CC.5,UCUT.4}. Also, CPGC outperforms both UC-MMC and MCC thanks to coded inputs. It also allows partial gradient computation, and provides approximately $25\%$ reduction in the average per iteration completion time compared to MCC and UC-MMC at a $5\%$ tolerance rate.


Communication volume of the UC-MMC scheme for $0\%$ tolerance rate is around $1.8$, which means that there is $80\%$ communication overhead. Similarly, the communication volume of CPGC is around $1.5$, which means a $50\%$ overhead. MCC has the minimum communication volume since the MDS code has zero decoding overhead\footnote{Communication volume of the MCC is slightly greater than $1$  since $K$ is not divisible by $r$, and zero padding is used before  encoding.}. We also observe that the communication volume of CPGC decreases with the tolerance level, and  it is close to that of MCC at a  tolerance level of around $10\%$.

We recall that the design goal of the CPGC scheme is to provide flexibility in seeking a balance between the per iteration completion time and accuracy. To this end, different iteration termination strategies can be introduced to reduce the overall convergence time. We show in \cite{long} that a faster overall convergence can be achieved with CPGC by increasing the tolerance at each iteration, as this would reduce the per-iteration completion time. Finally, one can observe from Fig. (\ref{comm1}) and (\ref{pack1}) that the MMC approach affects the communication load more drastically compared to the communication volume. This may introduce additional delays depending on the computing infrastructure and the communication protocol employed, e.g., dedicated links from the workers to the master compared to a shared communication network. 



%
%
%


\vfill\pagebreak

\bibliographystyle{IEEEbib}
\bibliography{strings,refs}

\end{document}